\documentclass[conference]{IEEEtran}
\IEEEoverridecommandlockouts
\usepackage{cite}
\usepackage{amsmath,amssymb,amsfonts}
\usepackage{algorithmic}
\usepackage{graphicx}
\usepackage{textcomp}
\usepackage{xcolor}
\usepackage{enumitem}
\usepackage{multirow}
\def\BibTeX{{\rm B\kern-.05em{\sc i\kern-.025em b}\kern-.08em
    T\kern-.1667em\lower.7ex\hbox{E}\kern-.125emX}}
    
\begin{document}

\title{Configuration Design of Mechanical Assemblies using an Estimation of Distribution Algorithm and Constraint Programming \\
}

\author{\IEEEauthorblockN{Hyunmin Cheong, Mehran Ebrahimi, Adrian Butscher, Francesco Iorio}
\IEEEauthorblockA{\textit{Autodesk Research} \\
Toronto, Canada \\
\{firstname.lastname\}@autodesk.com}
}

\maketitle

\begin{abstract}
A configuration design problem in mechanical engineering involves finding an optimal assembly of components and joints that realizes some desired performance criteria. Such a problem is a discrete, constrained, and black-box optimization problem. A novel method is developed to solve the problem by applying Bivariate Marginal Distribution Algorithm (BMDA) and constraint programming (CP). BMDA is a type of Estimation of Distribution Algorithm (EDA) that exploits the dependency knowledge learned between design variables without requiring too many fitness evaluations, which tend to be expensive for the current application. BMDA is extended with adaptive chi-square testing to identify dependencies and Gibbs sampling to generate new solutions. Also, repair operations based on CP are used to deal with infeasible solutions found during search. The method is applied to a vehicle suspension design problem and is found to be more effective in converging to good solutions than a genetic algorithm and other EDAs. These contributions are significant steps towards solving the difficult problem of configuration design in mechanical engineering with evolutionary computation.
\end{abstract}


\section{Introduction}
A notable trend in engineering is the increasing application of artificial intelligence and computational techniques to automate and improve design and manufacturing workflows. In particular, improving the early-stage design process can lead to significant cost-savings associated with the development and manufacturing of optimally functioning products \cite{suh}. 

\begin{figure}[t]
\centerline{\includegraphics[width=\linewidth]{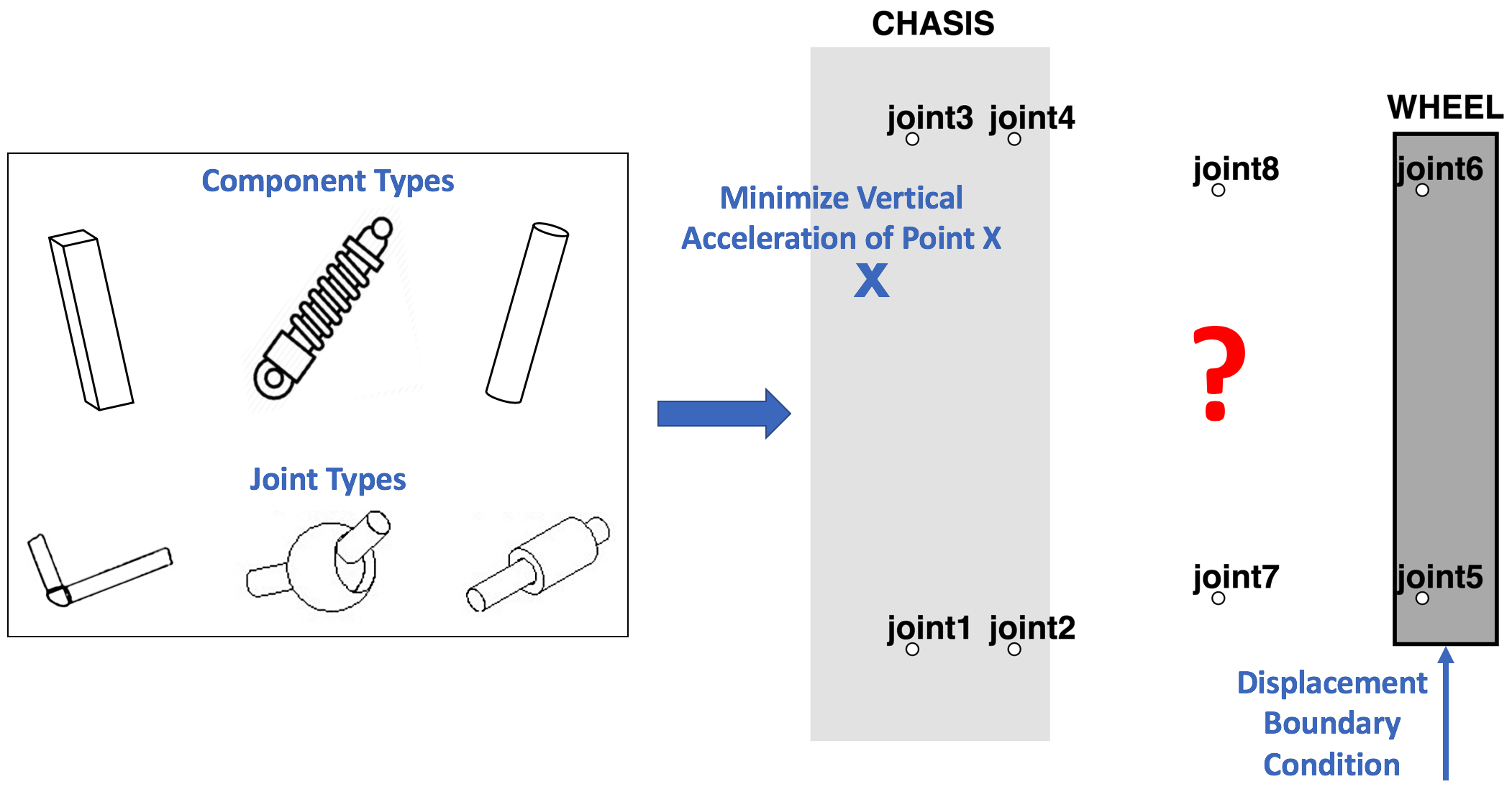}}
\caption{Configuration design problem for a mechanical assembly.}
\label{config}
\end{figure}

One challenging early-stage design problem in mechanical engineering is configuration design \cite{mittal}, defined as follows. Given a fixed set of predefined components and joints, find the optimal combination of components interfaced via joints that exhibits the desired mechanical behavior. An example of such a problem would be to compose a suspension system out of beams, springs, and dampers via fixed or spherical joint types, that would have minimal acceleration at a particular point on the chassis, as illustrated in Fig.~\ref{config}.

The configuration design problem can be characterized as a discrete, constrained, and black-box optimization problem, which is difficult to solve. First, we can treat the configuration design problem as operating in the space of graphs, where a particular graph represents a mechanical assembly with vertices as joints and edges as components. Each node and edge can be considered as a categorical design variable whose values specify the types of components and joints being used in the assembly. Also, several constraints must be satisfied for a given configuration to be considered feasible as a physically realizable and practical mechanical assembly. Moreover, the precise analytical form of the fitness function for a given solution depends on its configuration. Hence, evaluation of each solution requires invoking a physics-based simulation solver with the configuration as input.

Considering the above characterization, the current work applies an evolutionary computation (EC) technique to solve configuration design problems. In particular, this paper investigates using an Estimation of Distribution Algorithm (EDA). EDAs are a class of evolutionary algorithms \cite{baluja, bonet, harik1, harik2, pelikan1, pelikan2, santana, lima, shakya1, shakya2, alden} that use probability distributions estimated from a pool of candidate solutions to sample new solutions at each iteration of optimization. Reasons for choosing the EDA approach are as follows. 

First, fitness evaluation is expensive for mechanical assemblies, especially for those exhibiting dynamic behaviors. For example, evaluating the fitness of a suspension system requires physics-based simulation of its behavior over a specified time period. Even with the most advanced solvers, it can take in the order of minutes to evaluate a single solution. Hence, it is critical to minimize fitness evaluations as much as possible. 

Second, most mechanical assemblies feature dependencies between their constituting components. Such dependencies should be leveraged in solving the problem, either by learning them during the optimization process or being specified by the designer. If the dependency knowledge is learned, it can be communicated to the designer for better interpretation of the solutions found and exploring alternatives. The knowledge can also be reused for solving similar design problems later.

An EDA is deemed suitable in addressing the above challenges. First, for various EDAs, it has been shown that a fewer number of function evaluations are required to find optimal solutions compared to genetic algorithms (GA) \cite{bonet, harik1, pelikan1, shakya1, shakya2, martins}. Second, many EDAs exploit the structure of the problem, e.g., dependencies between variables, to generate new solutions at each iteration of the optimization process \cite{bonet, harik2, pelikan1, pelikan2, santana, lima, shakya1, shakya2, alden}. If the designer has prior knowledge about the problem, this can be introduced in the form of a probability distribution model. Also, most EDAs can learn the problem knowledge as part of optimization, which can be communicated to the designers for their benefit.

The particular EDA applied for the current work is Bivariate Marginal Distribution Algorithm (BMDA) \cite{pelikan1}. BMDA assumes bivariate dependencies between pairs of variables and uses the dependency statistics to compute the conditional probabilities of variables, which are then used to sample new solutions. The current work extends BMDA with adaptive chi-square testing for probability model estimation and Gibbs sampling for new population generation. The reasons for choosing and extending BMDA are provided in Related Work.

Lastly, an important aspect of our method is the use of constraint programming (CP) to handle configuration constraints. A mechanical assembly must satisfy a number of configuration constraints to be considered mechanically feasible. For example, all components must be connected to the assembly, which is equivalent to the connectivity condition in graph theory. Also, some types of components should only be associated with specific types of joints. Such constraints cannot be expressed only using algebraic operators and the degree of constraint violation cannot be quantified. Also, assemblies that violate the constraints sometimes cannot be evaluated because a physics-based solver may fail to simulate infeasible configurations. Therefore, applying many of the common constraint handling methods \cite{coello} is not appropriate. Because of this issue, our method uses CP to repair each candidate solution before evaluation. This process can be thought as performing local search to find a nearby solution that satisfies all the configuration constraints.

In summary, the main contributions of the paper are: 1) novel application of EDA and CP on a discrete, constrained, and black-box optimization problem, 2) extension of BMDA with adaptive chi-square testing and Gibbs sampling, and 3) a CP model to enforce the feasibility of mechanical assemblies.

The rest of the paper is organized as follows. Related Work presents prior applications of EC for engineering design, different types of EDAs considered, and the reasons behind appying BMDA for the current work. Next presented are the formulation of the configuration design problem and the optimization method. An experiment with an application example has been conducted to demonstrate the benefits of our method. Finally, the paper ends with Summary and Conclusions.

\section{Related Work}

\subsection{Related Applications of EC for Engineering Design}

Evolutionary computing techniques have been applied to solve various engineering design problems. Most closely related to the current work are those tackling the configuration design of engineering systems. In particular, GAs have been applied to find optimal systems from a discrete set of components \cite{angelov, grignon} similar to the current work. In both studies, however, it was relatively inexpensive to evaluate fitness functions because their analytical forms as a function of design variables were available, which is not the case for us. Also, in the prior work, the degrees of constraint violation could be quantified and incorporated into the objective function as penalties, which is also not possible for the current work. 

A number of studies have applied different EC techniques, such as GAs \cite{chapman, wu, kita, guirguis}, covariance matrix adaptation \cite{aulig}, and particle swarm optimization \cite{islam}, to solve topology or structural optimization problems. In such studies, continuum mechanical structures are discretized into finite cells and the problem is treated as a discrete optimization problem. Common to all these studies is that the design variables are binary and represent whether or not a material exists in a particular cell. In contrast, the current work deals with multiple categorical values for design variables, representing different types of components and joints. Related to this difference, the types of constraints a solution must satisfy to guarantee a feasible structure in the current work are more complex than the prior work. Lastly, the fitness evaluation of dynamic systems as in the current work requires significantly more time than static structures dealt in the prior work.

\subsection{Estimation of Distribution Algorithms}

Several types of EDAs have been considered for the current work. The earliest methods such as Population-Based Incremental Learning (PBIL) \cite{baluja} and Compact Genetic Algorithm (CGA) \cite{harik1} use univariate marginal frequencies as the probability model, which can be simple to compute but do not capture the dependencies between variables. Other notable methods such as Mutual-Information-Maximizing Input Clustering \cite{bonet}, Extended Compact Genetic Algorithm \cite{harik2}, Bivariate Marginal Distribution Algorithm (BMDA) \cite{pelikan1}, and Bayesian Optimization Algorithm \cite{pelikan2} leverage dependencies between pairs of variables and use conditional probability models. Pushing this idea further, a number of methods have been developed to use Markov Random Fields (MRF) as the dependency structure found among variables, considering either the global Markov property \cite{santana, shakya1, alden} or the local Markov property \cite{shakya2} in building the probability model and sampling a new population. 

\subsection{Justification for the Algorithm Choice}

The reasons for choosing BMDA for the current work is as follows. An intrinsic trade-off between the required number of function evaluations and the complexity of a probabilistic model used in an EDA is assumed. An EDA that leverages a complex probabilistic model, such as MRFs, can learn more sophisticated problem knowledge that can be more helpful in finding optimal solutions. However, learning such complex models would require a larger population size and hence a greater number of function evaluations. On the other hand, a simple EDA such as PBIL or CGA does not estimate the dependencies between variables and hence does not exploit any dependency knowledge. BMDA is deemed to strike the right balance -- it learns the pairwise dependencies between variables, which is not as sophisticated as MRF-based EDAs but still exploits the dependency knowledge.

Another important criterion is that BMDA requires less parameter tuning than more sophisticated methods. Our goal is to develop a method that can be easily applied to different design problems, so minimizing the number of algorithm parameters is one of the priorities. For example, Markovianity-based Optimization Algorithm (MOA) \cite{shakya2} requires one to set multiple parameters such as cross entropy threshold and temperature cooling rate that need to be tuned for the problem, and it was unclear how to set these appropriately for the current work. In fact, the numerical experiment presented later will demonstrate how MOA with the reported parameters in \cite{shakya2} did not work for our application example. On the other hand, the important parameter that needs to be set for BMDA, which is the chi-square test threshold value, can be intuitively determined based on its well-known statistical theory.

In the original BMDA \cite{pelikan1}, sampling new solutions relies on constructing a dependency graph between variables. Some variables are randomly selected to be independent, i.e., they become root nodes in the dependency graph. The variable values can then be sampled in a deterministic manner, first with the independent variables using their marginal probabilities, followed by the dependent variables using the conditional probabilities in the order of dependencies. To avoid the arbitrary selection of independent variables, the current work uses Gibbs sampling inspired from MRF-based EDA methods \cite{santana, shakya1, shakya2, alden}, to sample new solutions instead of constructing a dependency graph. While Gibbs sampling used in the prior work based on MRF distribution models required tuning of the temperature parameter \cite{shakya1, shakya2}, Gibbs sampling used for the current work is parameter-free because our method simply uses the conditional probabilities found between pairs of variables. Hereinafter, our chosen method of BMDA with Gibbs sampling is referred as BMDA-GS.

\section{Problem Formulation}

The configuration design problem of mechanical assemblies is formally presented. The formulation adheres to the following convention: An uppercase letter indicates a variable, a lowercase letter indicates a value assigned to a variable, and a bold letter indicates a set of variables or a collection of values.

A mechanical assembly can be represented as an undirected graph $\mathbf{G}=\langle \mathbf{Y}, \mathbf{Z} \rangle$, with its vertices $\mathbf{Y}$ as possible joints and edges $\mathbf{Z}$ as possible components associated with those joints. Indices $i$ and $j$ for an edge (component) indicate which vertices (joints) the edge is associated with. For example, $Z_{1,2}$ indicates a potential component that exists between joints $Y_1$ and $Y_2$. Furthermore, vertices and edges are treated as discrete variables, whose values $\mathbf{y}$ and $\mathbf{z}$ are defined over domains $\mathbf{D}^{(y_i)}$ and $\mathbf{D}^{(z_i)}$. For both variables, $0$ indicates that the corresponding joint or component is not used in the assembly, while non-$0$ values indicate different types of joints or components being used. For a given problem, $N$, $V$, and $W$ define the maximum number of joints, the number of joint types, and the number of component types, respectively. Note that $z_{i,j} = z_{j,i}$ and $z_{i,i} = 0$ since $\mathbf{G}$ is an undirected graph. 
\begin{gather}
\mathbf{Y} = \{Y_i\}, \; i = 1,...,N \label{eq:1} \\
\mathbf{Z} = \{Z_{i,j}\}, \;  i,j = 1,...,N \label{eq:2} \\
\mathbf{y} = \{y_i|y_i \in \mathbf{D}^{(y_i)}\}, \; \mathbf{D}^{(y_i)} = (0, 1, ..., V) \label{eq:3} \\
\mathbf{z} = \{z_{i,j}|z_{i,j} \in \mathbf{D}^{(z_{i,j})}\}, \; \mathbf{D}^{(z_{i,j})} = (0, 1, ..., W) \label{eq:4}
\end{gather}

Next, the notion of an \emph{environment object} needs to be introduced. An environment object, $envo$, can be thought as the mechanical assembly's link to the outside world, e.g., a wheel or a chassis for a suspension system. Some subset of joints in the assembly must belong to an environment object, i.e., $\langle \mathbf{\overline{Y}}_m, envo_m \rangle$. $M$ number of environment objects, $\textbf{envo}$, and joint-environment object membership pairs, $\textbf{joint\_envo\_pairs}$, are problems-specific definitions provided by the designer.
\begin{gather}
\textbf{envo} = \{envo_m\}, \; m = 1,...,M \label{eq:5} \\
\textbf{joint\_envo\_pairs} = \{\langle \mathbf{\overline{Y}}_m, envo_m \rangle\}, \;
\mathbf{\overline{Y}}_m \subset \mathbf{Y} \label{eq:6}
\end{gather}

We assume that a component cannot exist between two joints that belong to the same environment object, meaning
\begin{gather}
z_{i,j} = 0, \; \text{if } (Y_i \in \mathbf{\overline{Y}}_m \text{ and } Y_j \in \mathbf{\overline{Y}}_m) \label{eq:7}
\end{gather}

We now introduce a compound variable that is useful for the remaining formulation. Let $\mathbf{X}$ be a union of $\mathbf{Y}$ and $\mathbf{Z}$, representing the entire set of design variables. $\mathbf{X}$ can also be considered as a vectorized representation of $\mathbf{G}$. Let $\mathbf{x}$ denote values assigned to each of the unionized variables. Therefore, a particular collection of values, e.g., $\mathbf{x}^0$, is one instantiation of the graph $\mathbf{G}$.
\begin{gather}
\mathbf{X} = \mathbf{Y} \cup \mathbf{Z} \label{eq:8} \\
\mathbf{x} = \{x_i|x_i \in \mathbf{D}^{(x_i)}\}, \; i = 1, ..., |\mathbf{Y}|+|\mathbf{Z}| \label{eq:9}
\end{gather}

The following optimization problem can now be posed. 
\begin{gather}
\begin{aligned}
& \underset{\mathbf{X}}{\text{minimize}}
& & f(\mathbf{X}) \\
& \text{subject to}
& & C_i(\textbf{x}) \rightarrow \top, \; i = 1,...,L \label{eq:10}
\end{aligned}
\end{gather}
Here, computing the fitness function $f(\mathbf{X})$ requires invoking a physics-based simulation solver and passing as input a multibody assembly constructed according to $\mathbf{X}$. The solver will also take user-specified input for the problem such as load cases and boundary conditions. 

In addition, a solution must satisfy $L$ number of configuration constraints $C_i(\mathbf{X})$, to be considered feasible. $C_i(\mathbf{X})$ takes on Boolean values depending on whether a particular constraint is satisfied or violated.

\section{Optimization Method}

To solve the configuration design problem, an optimization method has been developed using BMDA-GS and CP-based repair operators. BMDA-GS is first presented, followed by the CP model created to perform repair operations.

\subsection{BMDA with Gibbs Sampling}

The following is the procedure of the BMDA-GS algorithm employed for the current work. Most of the procedure is based on the original BMDA \cite{pelikan1}. There are, however, two major differences inspired by other prior work on EDAs. First, for estimation of probability distributions, we consider the degrees of freedom observed from a selected population when applying the chi-square statistics, as in \cite{alden}. Also, for new population generation, we use Gibbs sampling instead of constructing a dependency graph between variables, as in \cite{santana, shakya1, shakya2}.

\subsubsection{Generation of Initial Population}

At the start of the algorithm, an initial population of solutions $\mathcal{P}$ is randomly generated. As a reminder, a solution is particular instantiation of the variables $\mathbf{x}$ in \eqref{eq:9}. If necessary, each solution is then repaired with the CP repair operators, which will be explained in the forthcoming subsection.

\subsubsection{Evaluation and Selection}

Solutions in the population are evaluated by invoking a physics-based solver and ranked based on their fitness values. A subset of the population, $\mathcal{S}$, representing top $T$ solutions is selected from $\mathcal{P}$. $|\mathcal{S}| / |\mathcal{P}|$ is defined as the truncation rate. 

\subsubsection{Estimation of Probability Distribution}

From $\mathcal{S}$, the probability distributions of the variables in the solutions are estimated. The probability model used by BMDA-GS is
\begin{gather}
p(\mathbf{x}) = \prod_{i=1}^{|\mathbf{X}|} p(x_i|\mathbf{N}_i), \; |\mathbf{N}_i|\leq 1  \label{eq:11}
\end{gather}
It is assumed that the probability of a variable $X_i$ can be computed with $p(x_i|\mathbf{N}_i)$, a conditional probability with dependency on its \emph{neighboring} set of variables, $\mathbf{N}_i$, which are the variables with the strongest dependency to $X_i$.

For the current work, $\mathbf{N}_i \leq 1$. Because fitness evaluation is expensive, the size of the population from which conditional probabilities can be obtained is limited. Therefore, computing conditional probabilities considering a large size of $\mathbf{N}_i$ is often impossible due to lack of required observations.

BMDA-GS by default does not assume any dependency between variables as prior knowledge. Such dependencies are learned during the optimization procedure by performing chi-square tests \cite{pelikan1}. A chi-square test is a statistical hypothesis test that identifies significant differences between the expected frequencies and the observed frequencies of categorical data from a given sample. If the differences are significant enough, it is assumed that there is a dependency between the categories involved. The chi-square test value for the sample $S$ between variables $x_i$ and $x_j$ is computed as follows
\begin{gather}
\chi_{i,j}^2 = \sum_{x_i,x_j} \frac{\big(|\mathcal{S}|p_{i,j}(x_i,x_j)-|\mathcal{S}|p_i(x_i)p_j(x_j)\big)^2}{|\mathcal{S}|p_i(x_i)p_j(x_j)} \label{eq:12}
\end{gather}
where $p_{i,j}(x_i,x_j)$, $p_i(x_i)$, and $p_j(x_j)$ are computed from $\mathcal{S}$.

Whether a particular chi-square test value is deemed significant or not depends on two factors. First, a confidence level must be chosen depending on the strictness of significance detection. For the current work, the confidence level of 99\% was found to be effective. Second, the significance depends on the degrees of freedom derivable from the observed sample. The degrees of freedom available for finding the dependency between $x_i$ and $x_j$ can be computed as follows \cite{alden}
\begin{gather}
\delta_{i,j} = (|\mathbf{D}^{(x_i)}|-1)(|\mathbf{D}^{(x_j)}|-1) - m_i - m_j \label{eq:13}
\end{gather}
where $m_i$ and $m_j$ are the numbers of missing observations for each value of variables $x_i$ and $x_j$, respectively, in the sample. For example, $m_i=2$ if two values of $x_i$ are never observed. 

Depending on the degrees of freedom available, the dependency between variables $x_i$ and $x_j$ is determined using the following criteria, based on the 99\% confidence level \cite{greenwood}:
\begin{gather}
\chi_{i,j}^2 \geq
\begin{cases} 
13.28, & \text{if } \delta_{i,j}=4 \\
11.34, & \text{if } \delta_{i,j}=3 \\
9.21, & \text{if } \delta_{i,j}=2 \\
6.63, & \text{if } \delta_{i,j}=1
\end{cases} \label{eq:14}
\end{gather}

Using the above chi-square test method, the neighboring set of dependent variables $\mathbf{N}_i$ is identified for each variable $x_i$. Then, we can compute the conditional probability $p(x_i|\mathbf{N}_i)$ using the frequency statistics of $\mathcal{S}$, considering the most dependent variable in $\mathbf{N}_i$, i.e., the variable with the highest chi-square test value. If $\mathbf{N}_i$ is empty for a particular variable, i.e., no statistically significant dependent variable is found, only its marginal probability $p(x_i)$ is computed from $\mathcal{S}$. 

\subsubsection{New Population Generation via Gibbs Sampling}

A new population is generated based on the probabilities computed from the previous generation of population and using Gibbs sampling. The generation procedure is performed as follows.

\begin{enumerate}[label=\roman*.]
    \item Initialize a solution $\mathbf{x} = \{x_1, x_2, ..., x_n\}$ at random.
    \item Randomly choose $x_i$ from $\mathbf{x}$.
    \item Update $x_i$ based on $p(x_i|\mathbf{N}_i)$ or $p(x_i)$.
    \item Repeat ii.-iii. for $1000*|\mathbf{x}|$ times.
    \item Take the final solution $\mathbf{x}$. 
\end{enumerate}

Note that the above procedure is used to generate a single solution. Hence, the procedure is repeated $|\mathcal{P}|$ times to generate a new population $\mathcal{P}'$. As in the case of initial population generation, each solution generated is repaired with the CP repair operators if necessary.

\subsubsection{Iterate Steps 2-4}

Steps 2 to 4 are repeated with the newly generated population until meeting a convergence criterion or an allocated number of iterations.

\subsection{Repair Operators based on Constraint Programming}

A solution that has been randomly generated or sampled from a probability distribution likely represents an infeasible configuration that cannot be simulated for evaluation. Therefore, we define a CP model that can be used to repair such a solution. Presented below are the CP model developed and how repair operations are performed using the model.

\subsubsection{Constraint Programming Model}

Constraints are imposed over the variables as defined in \eqref{eq:1} and \eqref{eq:2}. The first set of constraints imposed is on the connectivity of the configuration, based on the simple encoding technique by \cite{brown}. This can be done by ensuring that a path exists between every pair of active joints. A direct path exists between two joints either via a component between them or if the joints belong to the same environment object.

Let $\mathbf{A}$ represent a three-dimensional matrix where the $k^{th}$ dimension encodes whether a path length of $k$ exists between two joints $Y_i$ and $Y_j$. The domain values of $a_{i,j,k}$ indicate whether a path exists (value 1) or not (value 0).
\begin{gather}
\mathbf{A} = \{a_{i,j,k}\}, \; i,j = 1,...,N \text{ and } k = 1,...,N-1 \label{eq:15} \\
a_{i,j,k} \in (0, 1) \label{eq:16}
\end{gather}
For $k=1$, which is the dimension that indicates the existence of a direct path between joints, $a_{i,j,1} = 1$ if the corresponding component value $z_{i,j}$ is non-0, i.e., there is a component between joints $Y_i$ and $Y_j$, or the joint pair belongs to the same environment object. 
\begin{gather}
\forall i,j: a_{i,j,1} = 1, \;\; \text{if } (z_{i,j} \geq 1) \text{ or } (Y_i, Y_j \in \mathbf{\overline{Y}}_m) \label{eq:17}
\end{gather}
Next, a path of $k$ length must exist between each pair of joints. 
\begin{gather}
\forall i,j \; (i<j):
\sum_{k=1}^{N-1} a_{i,j,k} > 0 \label{eq:18}
\end{gather}
Lastly, it is enforced that a path of length $k$ exists between joints $Y_i$ and $Y_j$, if there exists a third joint that is directly connected to $Y_i$ and has a path of length $k-1$ to $Y_j$.
\begin{gather}
\forall i,j \; (i<j) \text{ and } k = 2,...,N-1: \nonumber \\
a_{i,j,k} = 1 , \;\; \text{if } (\vec{v_1} \cdot \vec{v_2} \neq 0) \label{eq:19}\\
\vec{v_1} = (a_{i,1,1}, a_{i,2,1},..., a_{i,N,1}) \label{eq:20}\\
\vec{v_2} = (a_{1,j,k-1}, a_{2,j,k-1},..., a_{N,j,k-1}) \label{eq:21}
\end{gather}
The above condition completes the connectivity constraint. 

Next, the following constraint is imposed: Joints that do not belong to an environment object must be associated with at least two components. This can be imposed by requiring the sum of all component values associated with each of such joints to be greater than 1.
\begin{gather}
\forall i,k:
\sum_{j}^{N} z_{i,j} > 1 \;\;\; \text{if } y_i \notin \{\mathbf{\overline{Y}}_k\} \label{eq:22}
\end{gather}

The last constraint represents problem-specific restrictions on joint types and component types that can be associated together. A number of such constraints can be applied where $g_c$ is the specific joint type that must be associated with the specific component type $h_c$. The special operator $[...]_\mathbb{Z}$ returns 1 if the expression inside is true or 0 otherwise. The constraint is equivalent to stating that $y_i$ must be $g_c$ if at least one of the components associated with it, $z_{i,j}$, is of the type $h_c$.
\begin{gather}
\forall i,c:
[y_i = g_c]_\mathbb{Z} = \big[\sum_j^N [z_{i,j} = h_c]_\mathbb{Z} > 0 \big]_\mathbb{Z}
\label{eq:23}
\end{gather}

\subsubsection{Performing Repair Operations}

The CP model is used to perform three repair operations in a sequence. For each operation, some of the variables in the CP model are instantiated based on the current solution to be repaired. Then, a generic CP solver is used to instantiate the values of the remaining variables such that all the constraints are satisfied. 

For the current work, the CP solver from Google's or-tools library (https://developers.google.com/optimization/) is used. The solver works by employing \emph{propagation} and \emph{backtracking} techniques. During propagation, a value is assigned to a randomly chosen variable. Then, values of other variables are eliminated from their domains using constraints to reduce the search space. The CP solver allows different value assignment strategies, e.g., {\footnotesize \textsc{ASSIGN\_MIN\_VALUE}}, {\footnotesize\textsc{ASSIGN\_MAX\_VALUE}}, {\footnotesize \textsc{ASSIGN\_RANDOM\_VALUE}}, etc. If the search gets stuck, i.e., when the solver cannot assign a value to the next variable without violating a constraint, the solver backtracks to the previous decision and assigns a different value to the variable.

\begin{figure}[b]
\centerline{\includegraphics[width=0.8\linewidth]{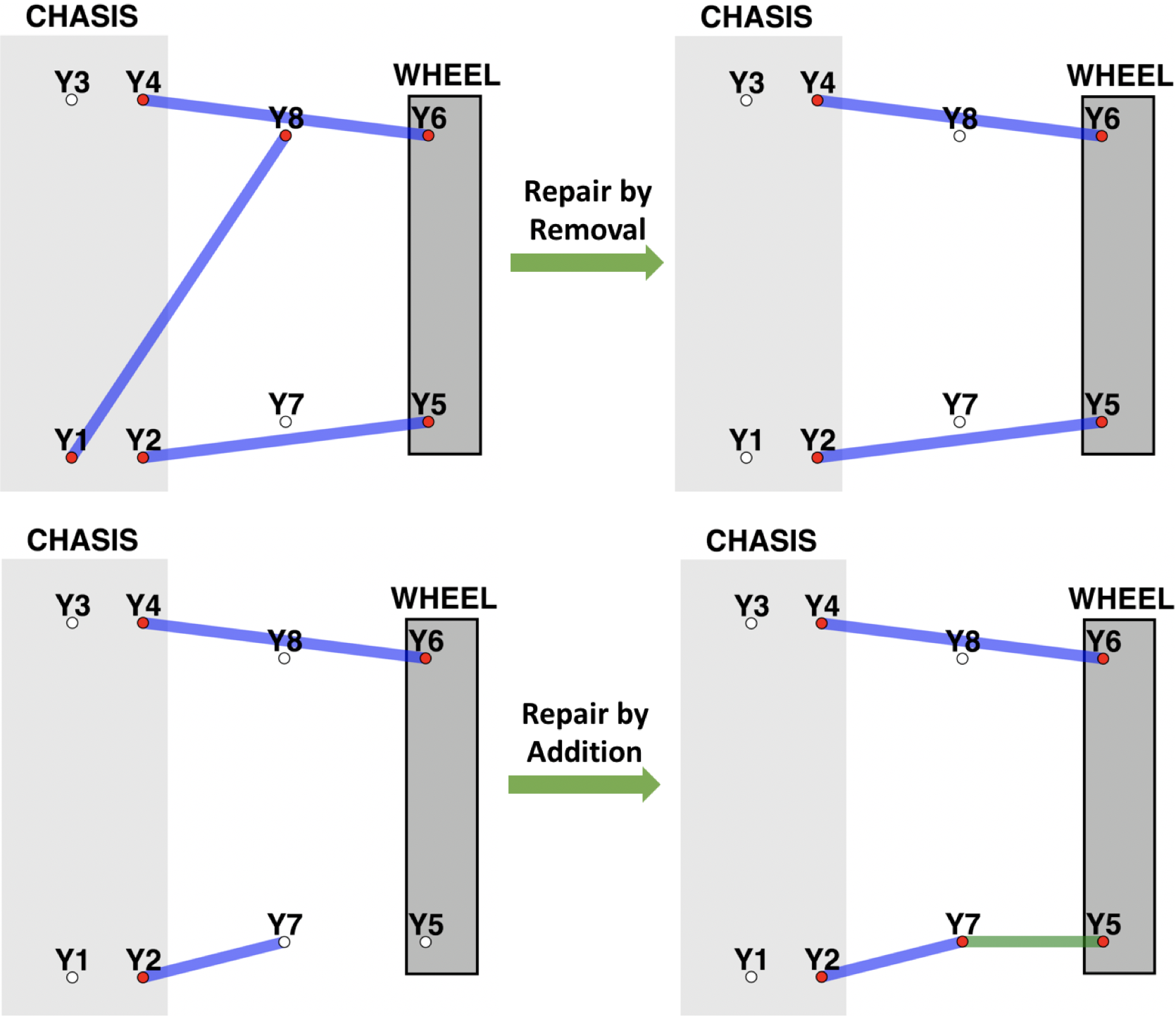}}
\caption{Repairing an infeasible configuration by removing a component (Top) and adding a component (Bottom).}
\label{repair}
\end{figure}

The first repair attempts to remove unnecessary components (Fig.~\ref{repair}, Top). In this case, all the component variables with 0 values in the current solution are fixed while the domains of other variables are set to 0 or its current value. Then, a search is performed with the {\small\textsc{ASSIGN\_MAX\_VALUE}} option, which attempts to assign a non-0 value to each variable until a constraint is violated. A feasible solution is found if the solution can assign non-0 values to all variables except for any variable that is assigned 0, while satisfying all the constraints.

If repair by removing a component is not possible, e.g., the configuration shown on the left in Fig.~\ref{repair} (Bottom), the second repair is done by adding components. For this repair, all the component variables with non-0 values in the current solution are fixed while the domains of other variables are set to 0 or a randomly chosen value among the non-0 component values. Then, a search is performed with the {\small\textsc{ASSIGN\_MIN\_VALUE}} option, which attempts to assign a 0 value to each variable except for any variable that is assigned a non-0 value to satisfy the constraints.

Lastly, joint variable values $y_i$ are repaired if necessary. The values are chosen such that the constraint \eqref{eq:23} is satisfied. 

The operations described above as a whole always guarantee that an infeasible configuration can be repaired into a feasible one during the optimization process. The results from the numerical experiment will show the importance of the CP repair operators in finding good solutions.

\section{Experiment with an Application Example}

\subsection{Application: Configuration Design of a Suspension System}

\begin{figure}[t]
\centerline{\includegraphics[width=0.8\linewidth]{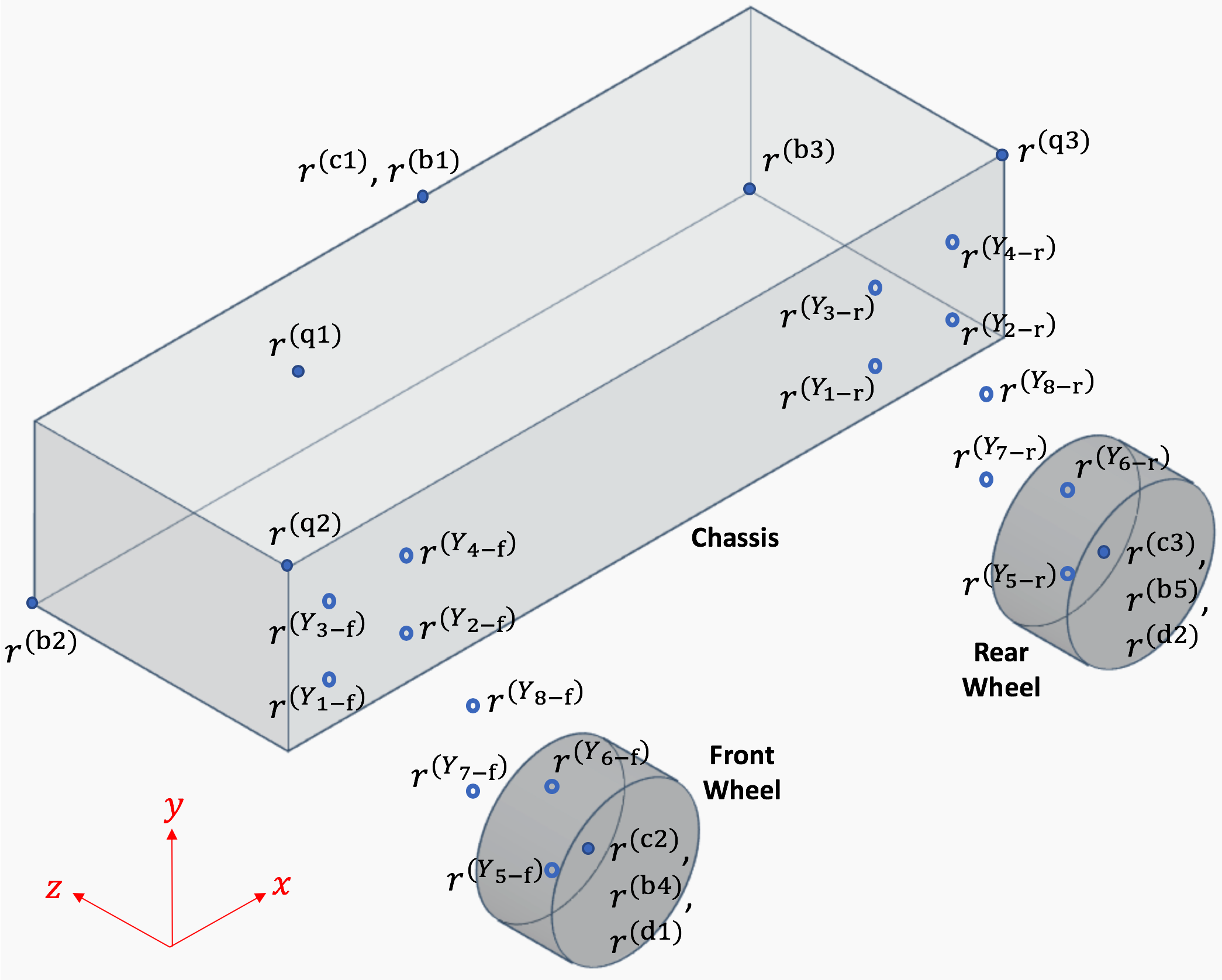}}
\caption{Illustration of the suspension design problem. Non-filled dots are variable joints. $r^{(\text{q[i]})}$ represent the points considered in the fitness function. $r^{(\text{c[i]})}$ represent the center-of-mass for each environment object. Boundary conditions are defined on $r^{(\text{b[i]})}$ and load conditions are defined on $r^{(\text{d[i]})}$.}
\label{prob}
\end{figure}

An experiment was conducted with an application example to demonstrate the benefits of the optimization method. Configuration design of a car suspension system was chosen as an example, as depicted in Fig.~\ref{prob}. A similar problem was tackled by \cite{shira, arik} using genetic algorithms, although in their cases the parameters of a fixed configuration were optimized. 

The following types of joints and components are considered. For joints: 0, 1, and 2 represent no joint, a welded joint, and a spherical joint, respectively. For components: 0, 1, and 2 indicate no component, a beam, and a shock absorber (a spring-damper combination), respectively. Each joint and component type exhibits different mechanical behaviors during simulation and therefore affects the fitness of a solution.

A problem-specific constraint per \eqref{eq:23} is imposed to constrain that a shock aborber can only be associated with a spherical joint: 
\begin{gather}
\forall i:
[y_i = 2]_\mathbb{Z} = \big[\sum_j^N [z_{i,j} = 2]_\mathbb{Z} > 0 \big]_\mathbb{Z}
\label{eq:24}
\end{gather}

The physical model considered is one half of the car, with boundary conditions applied along the dividing plane to mimic full-car simulation. Two distinct displacement boundary conditions are applied on the front and rear wheels, mimicking the car being driven on a bumpy terrain. An identical suspension configuration found by the optimization algorithm is used for both the front and back of the car during simulation.

Referring back to \eqref{eq:1}, the number of possible joints in the configuration are defined as $N=8$. Joints 1-4 belong to the chassis and Joints 5-6 belong to the wheel. Joints 7-8 are free joints that do not belong to any environmental object. After imposing the value constraints described in Section III, the number of component variables is determined to be 21. Hence, the total number of variables is $|\mathbf{X}|=29$. Note that a double wishbone suspension design, which is one of the most complex designs used in high performance vehicles, also consists of eight joints. Therefore, our experiment problem resembles the same degree of complexity as real-world suspension designs.

Table~\ref{tab1} lists other parameters of the problem required by the physics-based solver to evaluate solutions. The solver was developed in-house by the current authors \cite{ebrahimi}.

\begin{table}
\caption{Experiment Settings for the Suspension Design Problem}
\begin{center}
\setlength\tabcolsep{3pt}
\begin{tabular}{|l|ll|}
\hline

& $r^{(Y_{1\text{-f}})}=(0.1, 0.4, 0.3)$$^{\mathrm{a}}$ &  $r^{(Y_{2\text{-f}})}=(0.3, 0.4, 0.3)$ \\
Joint Positions& $r^{(Y_{3\text{-f}})}=(0.1, 0.6, 0.3)$ & $r^{(Y_{4\text{-f}})}=(0.3, 0.6, 0.3)$  \\
on Front Side& $r^{(Y_{5\text{-f}})}=(0.2, 0.3, 0.05)$ & $r^{(Y_{6\text{-f}})}=(0.2, 0.5, 0.05)$  \\
& $r^{(Y_{7\text{-f}})}=(0.2, 0.35, 0.15)$ & $r^{(Y_{8\text{-f}})}=(0.2, 0.55, 0.15)$  \\ \hline

& $r^{(Y_{1\text{-r}})}=(2.38, 0.4, 0.3)$ &  $r^{(Y_{2\text{-r}})}=(2.38, 0.6, 0.3)$ \\
Joint Positions  & $r^{(Y_{3\text{-r}})}=(2.58, 0.4, 0.3)$ & $r^{(Y_{4\text{-r}})}=(2.58, 0.6, 0.3)$  \\
on Rear Side & $r^{(Y_{5\text{-r}})}=(2.48, 0.3, 0.05)$ & $r^{(Y_{6\text{-r}})}=(2.48, 0.5, 0.05)$  \\
& $r^{(Y_{7\text{-r}})}=(2.48, 0.35, 0.15)$ & $r^{(Y_{8\text{-r}})}=(2.48, 0.55, 0.15)$  \\ \hline

\multirow{2}{*}
{Shock absorber} & \multicolumn{2}{l|}{spring constant $= 75000$ N/m} \\ 
& initial length $= 0.5$ m & damping ratio $= 875$ \\ \hline

\multirow{2}{*}
{Beam} & cross-x area $= 0.0007$ m$^2$ & density $= 8000$ kg/m$^3$ \\ 
 & elastic modulus $= 200$ GPa & shear modulus $= 70$ GPa \\
 & \multicolumn{2}{l|}{second moment of inertia, y = $3.3 \times 10^{-7}$ m$^4$} \\
 & \multicolumn{2}{l|}{second moment of inertia, z = $3.3 \times 10^{-7}$ m$^4$} \\
 & \multicolumn{2}{l|}{polar moment of inertia = $1.6 \times 10^{-7}$ m$^4$} \\
\hline

\multirow{5}{*}
{Chassis} & \multicolumn{2}{l|}{moment of inertia $= (48.2, 647, 694)$ kg$\cdot$m$^2$} \\
&  mass $= 1000$ kg & $r^{(\text{c1})} = (1.524, 0.6, 1.025)$  \\ 
& $r^{(\text{q1})} = (1.29, 0.7, 0.65)$ & {$r^{(\text{q2})} = (0, 0.6, 0.3)$} \\ 
& {$r^{(\text{q3})} = (2.68, 0.6, 0.3)$} & {$r^{\text(b1)} = (1.524, 0.6, 1.025)$} \\ & {$r^{\text(b2)} = (0, 0.4, 1.025)$} & {$r^{\text(b3)} = (2.68, 0.4, 1.025)$} \\ 
\hline

\multirow{3}{*}
{Front Wheel} & \multicolumn{2}{l|}{moment of inertia $= (0.686, 0.686, 0.972)$ kg$\cdot$m$^2$} \\
& mass $= 60$ kg & $r^{(\text{c2})} = (0.2, 0.4, 0)$ \\ 
& $r^{(\text{b4})} = (0.2, 0.4, 0)$  & $r^{(\text{d1})} = (0.2, 0.4, 0)$  \\ \hline

\multirow{3}{*}
{Rear Wheel} & 
\multicolumn{2}{l|}{moment of inertia $= (0.686, 0.686, 0.972)$ kg$\cdot$m$^2$} \\
& mass $= 60$ kg & $r^{(\text{c3})} = (2.48, 0.4, 0)$ \\ 
& $r^{(\text{b5})} = (2.48, 0.4, 0)$  & $r^{(\text{d2})} = (2.48, 0.4, 0)$ \\ \hline

 & \multicolumn{2}{l|}{$d_z = 0 \text{ at } r^{(\text{b1})}$ \; $d_z = 0 \text{ at } r^{(\text{b2})}$ \; $d_z = 0 \text{ at } r^{(\text{b3})}$} \\
Boundary & \multicolumn{2}{l|}{$d_x = 0 \text{ at } r^{(\text{b4})}$ \; $d_x = 0 \text{ at } r^{(\text{b5})}$}  \\ 
Conditions & \multicolumn{2}{l|}{$d_z = 0.050 \sin (2 \pi t)$ at $r^{(\text{d1})} = (0.2, 0.4, 0)$} \\
& \multicolumn{2}{l|}{$d_z=0.075 \sin (4 \pi t)$ at $r^{(\text{d2})} = (2.48, 0.4, 0)$} \\ \hline
\multicolumn{3}{l}{$^{\mathrm{a}}$Units for all position vectors, $r^{(x)}$, are in m.}
\end{tabular}
\label{tab1}
\end{center}
\end{table}

The fitness function for the example problem is a single composite function consisting of multiple objective criteria. 
\begin{equation}
f(\mathbf{X}) = \sum_t^T \big(a_{z,t}^{(\text{q1})} + w_1*(d_t^{(\text{q2})} + d_t^{(\text{q3})})\big) + w_2*|\mathbf{Y}_{y_i\neq0}| \label{eq:25}
\end{equation}
where $a_{z,t}^{(\text{q1})}$ is the vertical acceleration of point q1, $d_t^{(\text{q2})}$ is the total displacement of point q2, and $d_t^{(\text{q3})}$ is the total displacement of point q3, all located on the chassis and measured at time $t$. These quantities are summed over the simulation time period, $T$. Point q1 denotes the location of the driver seat, while points q2 and q3 denote the front and rear corners of the chassis. The last term $|\mathbf{Y}_{y_i\neq0}|$ represents the number of components used in the assembly. Weight factors $w_1$ and $w_2$ are applied to the displacement terms and the component number term, respectively. For the experiment, we set $w_1=5000$ and $w_2=100$. In summary, the objective is to minimize the vertical acceleration experienced by the driver, the total displacements at the corners of chassis that are related to the roll and the pitch of the chassis, and the number of components required in the assembly.

The complete solving procedure should involve bi-level optimization as in \cite{islam}, i.e., it should find an optimal solution considering both the configuration and its parameters such as joint positions and spring constants. In the current paper, only discrete variables are considered. Although we have developed a gradient-based optimization method for continuous parameter optimization \cite{ebrahimi}, it was not included in the experiment because it would take a considerable amount of time for fitness evaluations. Including such a capability would be equivalent to replacing one black-box objective function with another.

\subsection{Algorithms Compared}

We compared BMDA-GS against the original BMDA \cite{pelikan1}, GA, and MOA \cite{shakya2}, all with the CP-based repair operators enabled. Also, we ran BMDA-GS without the repair operators to examine the effect of using them. Considering the expensive evaluation process, the population size of 100 was used for all. For BMDA-GS, BMDA, and MOA, truncation rate $= 0.2$ was used. For BMDA-GS and BMDA, the $99\%$ chi-square test threshold was used. For GA, based on the prior work on suspension design \cite{arik} and selection method experiment \cite{zhang}, crossover rate $= 0.9$, mutation rate $= 0.1$, and binary tournament selection with truncation rate $= 0.6$ were used. For MOA, the parameters were set per \cite{shakya2}. For all algorithms, the averages of the best fitness values from 30 runs are reported.

All algorithms were run on a single desktop computer with an Intel Xeon Processor E5-2650 at 2.60GHz with 8 cores and 16 threads. This allowed 16 fitness evaluations of solutions in parallel, each of which taking about two minutes. On average, the overall optimization algorithm took about four hours.

\subsection{Results}

\begin{figure}[!b]
\centerline{\includegraphics[width=0.9\linewidth]{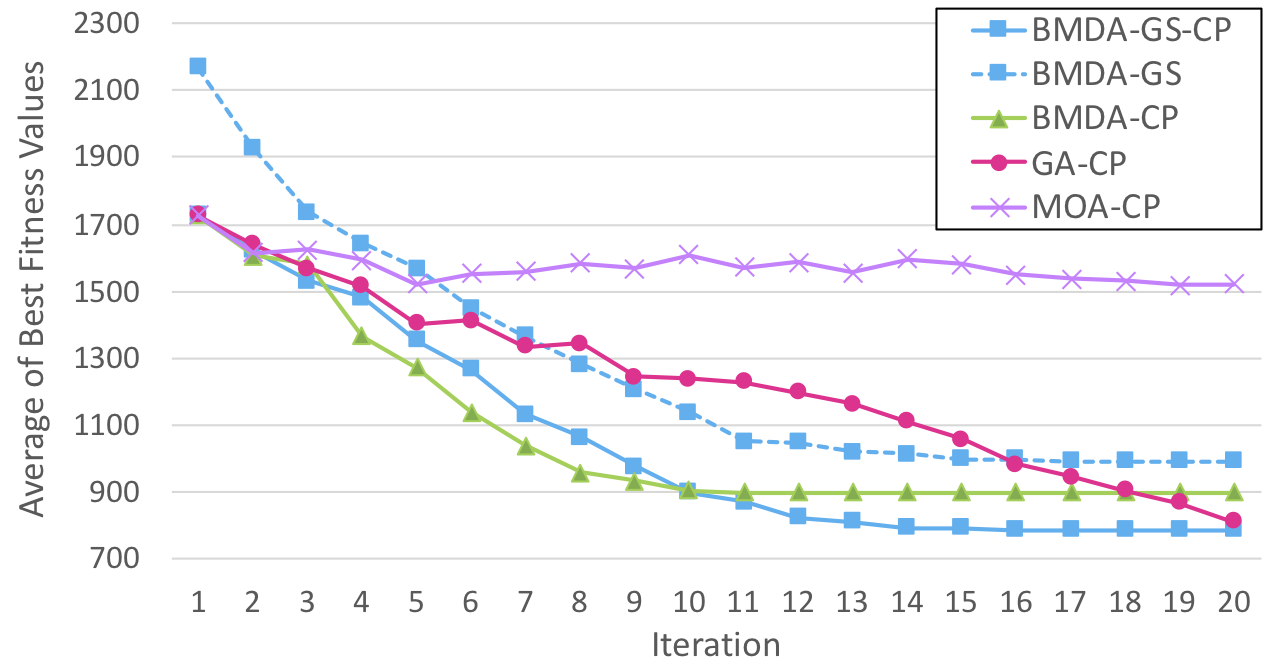}}
\caption{Comparison of algorithms.}
\label{exp1}
\end{figure}

Figure~\ref{exp1} shows the results. Algorithms combined with the CP-based repair operators have a trailing label ``-CP''. BMDA-GS-CP and BMDA-CP converged to good solutions faster than GA. MOA did not work at all. BMDA-GS-CP converged to a better solution than BMDA-CP, demonstrating the benefit of using Gibbs sampling. Although GA eventually reaches to a solution as good as BMDA-GS-CP, it took almost seven more iterations. Considering that fitness evaluation is expensive, such slow convergence is not desired. Also, the results of BMDA-GS-CP vs. BMDA-GS show that the repair operators helped in finding a better solution at each iteration. This is mainly because the repair operators can remove unnecessary components that otherwise penalize the fitness value. 

\begin{figure}[!b]
\centerline{\includegraphics[width=0.65\linewidth]{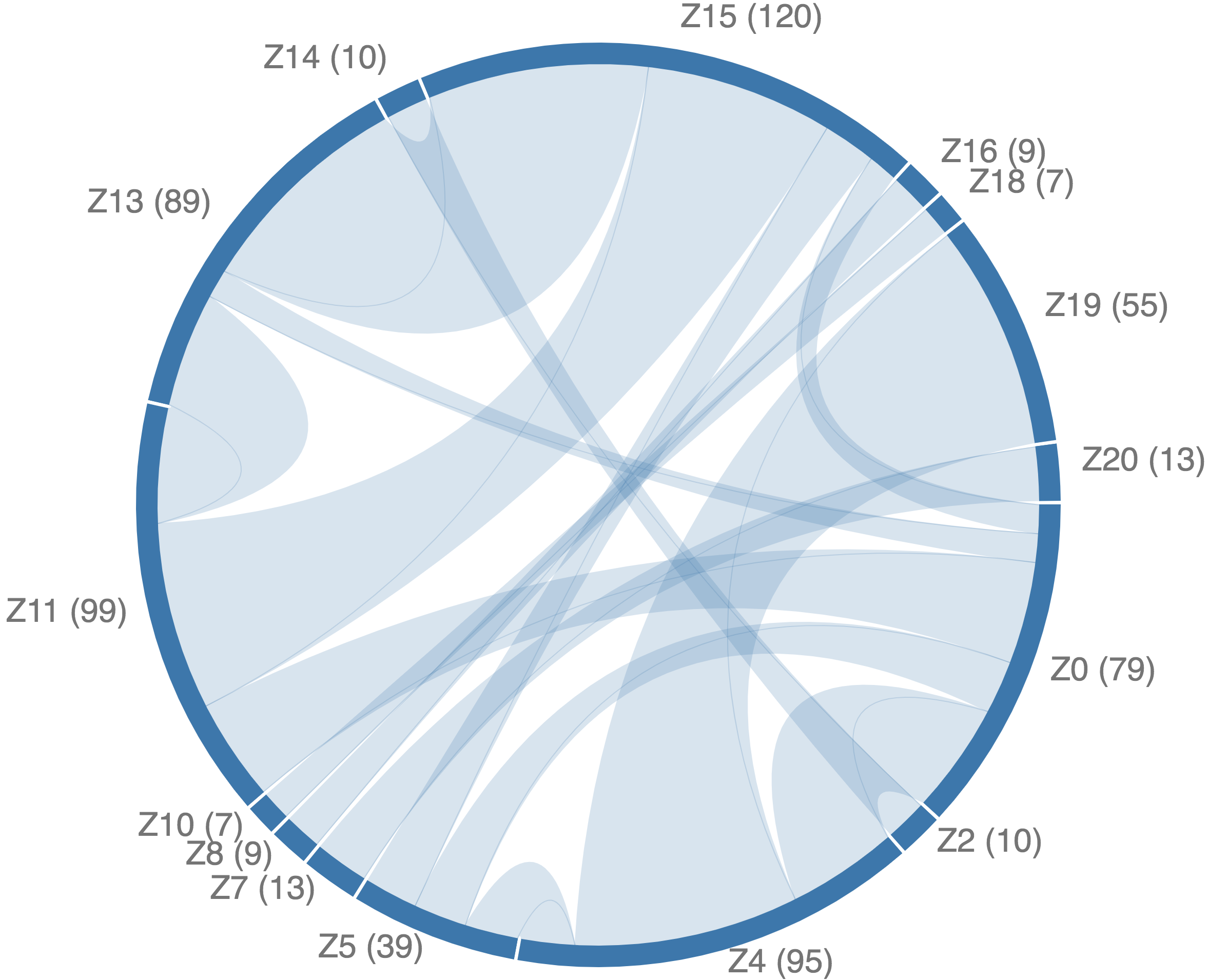}}
\caption{Visualization of dependencies between component variables. The numbers inside the parentheses are cumulative chi-square test values. }
\label{dep}
\end{figure}

As stated earlier, one benefit of EDA is that the knowledge gained during optimization could be communicated to the designer. This is important as designers in practice often would like to explore alternative solutions even after an optimal solution is given by an algorithm \cite{gaur}. Figure~\ref{dep} visualizes the dependencies between components, which were determined using chi-square tests as part of BMDA-GS-CP. The thickness of the chord connecting each component pair correlates with the cumulative chi-square test values obtained during optimization and indicates the strength of the dependency. If the designer decides to add or remove a component to a given assembly, the knowledge visualized can be referred to estimate the potential effect of that component on other components.

\begin{figure*}
\includegraphics[width=\textwidth]{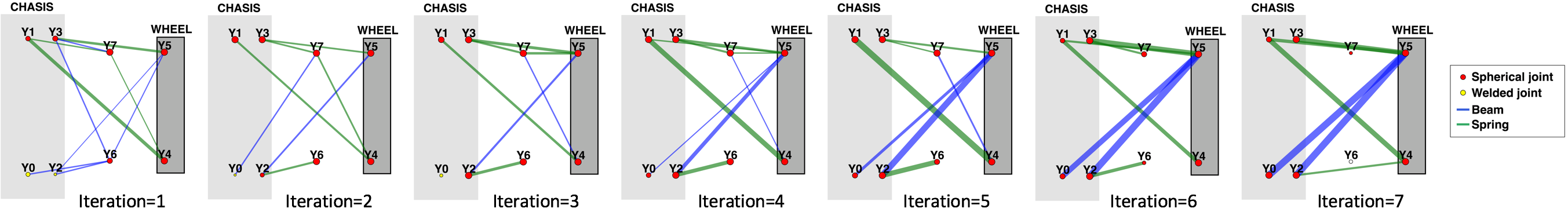}
\caption{Visualization of probabilistic distributions for each component and joint variable at different iterations. A flattened 2D view is shown.}
\label{evo}
\end{figure*}

In addition, Figure~\ref{evo} visualizes the evolution of probabilistic distributions for each component and joint variable obtained from BMDA-GS-CP. Different colors used for each component or joint indicate its type with the highest probability value and the size of the component or joint is correlated with its highest probability value. Instead of displaying the best solution at each iteration, this approach allows the designer to estimate the degree of effect that a particular component or joint has on the behavior of the assembly. For instance, the components with high probability values can be kept while those with lower probability values can be considered for modification.  

\section{Summary and Conclusions}

The current work demonstrates the effectiveness of using BMDA, a type of EDA, and CP to find an optimal configuration of a mechanical assembly, which can be characterized as a discrete, constrained, and black-box optimization problem. BMDA is applied because of its balance between exploiting a complex probability distribution model and the fitness evaluations required to learn such a model. Our method extends BMDA by incorporating adaptive chi-square testing and Gibbs sampling. In addition, the method features a CP model that is used to repair infeasible configurations generated during search into feasible ones. For the application of designing a car suspension system, both the extension of BMDA and the CP-based repair operators have been shown to have benefits on more effectively converging to good solutions.

Future work includes integrating the method developed as part of complete bi-level optimization, which would also consider the optimization of parametric variables for each configuration. In addition, we aim to consider different probability factorization methods for dealing with design problems with a larger number of design variables. Finally, the method developed could be applied to other similar optimization problems found in different domains, with the contributions of the current work serving as an important foundation.

\end{document}